\documentclass[twocolumn,10pt]{article}
\usepackage[T1]{fontenc}    
\usepackage{hyperref}       
\usepackage{url}            
\usepackage{booktabs}       
\usepackage{amsfonts}       
\usepackage{nicefrac}       
\usepackage{microtype}      
\usepackage{lipsum}

\usepackage{times}
\usepackage{graphicx}
\usepackage{subcaption}
\usepackage{float}
\usepackage{authblk}
\usepackage{amsmath}
\usepackage{amssymb}
\usepackage{epsfig}
\hypersetup{hidelinks=true}

\title{A Detection and Segmentation Architecture for Skin Lesion Segmentation on Dermoscopy Images}
\author{Chengyao Qian}
\author{Ting Liu}
\author{Hao Jiang}
\author{Zhe Wang}
\author{Pengfei Wang}
\author{Mingxin Guan}
\author{Biao Sun}
\affil{MTlab (Meitu, Inc.)}
\affil{\{qcy1, lt, jh1, wz, wpf1, gmx, sunbiao\}\textit @meitu.com}
\date{}

\begin{document}
{\centering
\maketitle

}

\begin{abstract}
This report summarises our method and validation results for the ISIC Challenge 2018 - Skin Lesion Analysis Towards Melanoma Detection - Task 1: Lesion Segmentation. We present a two-stage method for lesion segmentation with optimised training method and ensemble post-process. Our method achieves state-of-the-art performance on lesion segmentation and we win the first place in ISIC 2018 task1.
\end{abstract}


\section{Introduction}
Melanoma is the most dangerous type of skin cancer which cause almost 60,000 deaths annually. In order to improve the efficiency, effectiveness, and accuracy of melanoma diagnosis, International Skin Imaging Collaboration (ISIC) provides over 2,000 dermoscopic images of various skin problems for lesion segmentation, disease classification and other relative research. 

Lesion segmentation aims to extract the lesion segmentation boundaries from dermoscopic images to assist exports in diagnosis. In recent years, U-Net, FCN and other deep learning methods are widely used for medical image segmentation. However, the existing algorithms have the restriction in lesion segmentation because of various appearance of lesion caused by the diversity of persons and different collection environment. FCN, U-Net and other one-stage segmentation methods are sensitive to the size of the lesion. Too large or too small size of lesion decreasing the accuracy of these one-stage methods. Two-stage method could reduce the negetive influence of diverse size of lesion. Mask R-CNN can be viewed as a two-stage method which has an outstanding performance in COCO. However, Mask R-CNN still has some brawbacks in lesion segmentation. Unlike clear boundary between different objects in COCO data, the boundary between lesion and healthy skin is vague in this challenge. The vague boundary reduces the accuracy of RPN part in Mask R-CNN which may cause the negative influence in following segmentation part. Furthermore, the low resolution of input image in the segmentation of Mask R-CNN also reduce the accuracy of segmentation.

In this report, we propose a method for lesion segmentation. Our method is a two-stage process including detection and segmentation. Detection part is used to detect the location of the lesion and crop the lesion from images. Following the detection, segmentation part segments the cropped image and predict the region of lesion. Furthermore, we also propose an optimised processed for cropping image. Instead of cropping the image by bounding box exactly, in training, we crop the image with a random expansion and contraction to increase robustness of neural networks. Finally, image augmentation and other ensemble methods are used in our method. Our method is based on deep convolutional networks and trained by the dataset provided by ISIC  \cite{Tschandl2018_HAM10000} \cite{DBLP:journals/corr/abs-1710-05006}. 

\section{Materials and Methods}

\begin{figure*}
 \centering
  \includegraphics[height=6cm,width=0.75\textwidth]{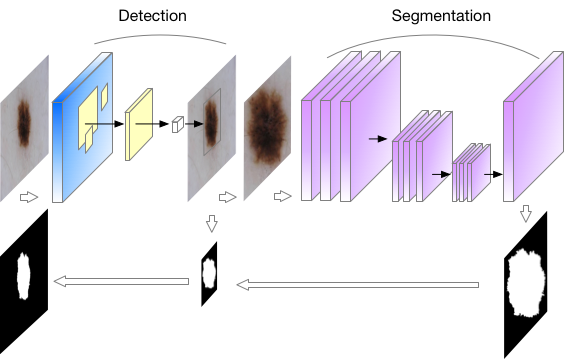}
  \caption{Process}
  \label{fig:Process}
\end{figure*}

\subsection{Database and Metrics}
For Lesion Segmentation, ISIC 2018 provides 2594 images with corresponding ground truth for training. There are 100 images for validation and 1000 images for testing without ground truth. The number of testing images in this year is 400 more than which in 2017. The size and aspect ratio of images are various. The lesion in images has different appearances and appear in different parts of people. There are three criterions for ground truth which are a fully-automated algorithm, a semi-automated flood-fill algorithm and manual polygon tracing. The ground truth labelled under different criterions has different shape of boundary which is a challenge in this task. We split the whole training set into two sets with ratio 10:1.

The Evaluation of this challenge is Jaccard index which is shown in Equation \ref{eq:jaccard}. In 2018, the organiser adds a penalty when the Jaccard index is below 0.65.
\begin{equation}
J(A,B) = 
\begin{cases}
\frac{|A \bigcap B|}{|A \bigcup B|}  & J(A,B) \geq 0.65 \\
0 & \text{otherwise}\\
\end{cases}
\label{eq:jaccard}
\end{equation}


\subsection{Methods}

We design a two-stage process combining detection and segmentation. Figure \ref{fig:Process} shows our process. Firstly, the detection part detects the location of the lesion with the highest probability. According to the bounding box, the lesion is cropped from the original image and the cropped image is normalised to 512*512 which is the size of the input image of segmentation part. A fine segmentation boundray of the cropped image is provided by the segmentation part.

\subsubsection{Detection Part}
In the detection process, we use the detection part in MaskR-CNN \cite{DBLP:journals/corr/HeGDG17}. We also use the segmentation branch of Mask R-CNN to supervise the training of neural networks.

\begin{figure}[H]
  \centering
  \includegraphics[width=0.4\textwidth]{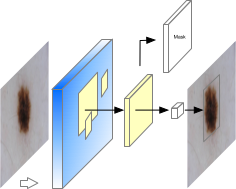}
  \caption{Detection}
  \label{fig:Detection}
\end{figure}

\subsubsection{Segmentation Part}
\begin{figure}[H]
  \centering
  \includegraphics[width=0.5\textwidth]{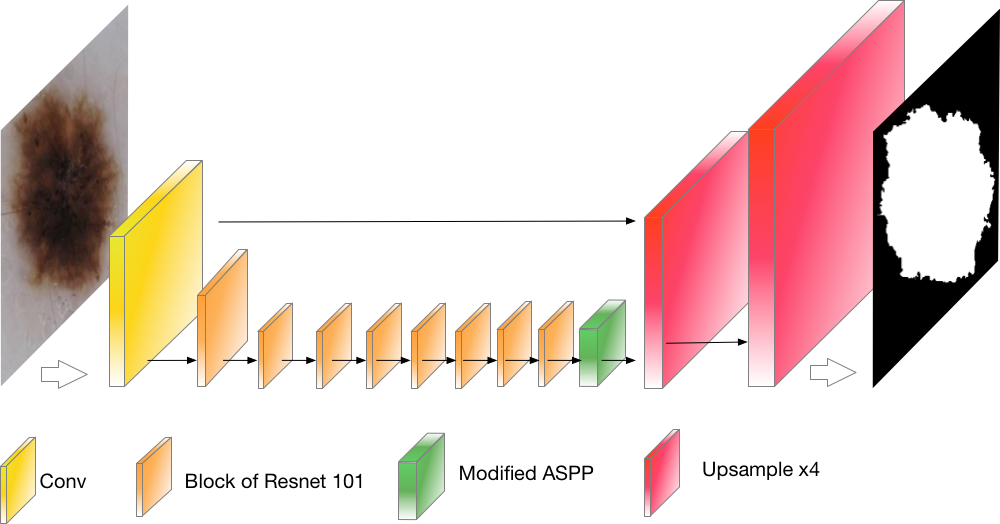}
  \caption{Segmentation}
  \label{fig:Segmentation}
\end{figure}

In the segmentation part, we design an encode-decode architecture of network inspired by DeepLab \cite{DBLP:journals/corr/ChenPK0Y16}, PSPNet \cite{DBLP:journals/corr/ZhaoSQWJ16}, DenseASPP \cite{Yang_2018_CVPR} and Context Contrasted Local \cite{Ding_2018_CVPR}. Our architecture is shown in Figure \ref{fig:Segmentation}. The features are extracted by extended ResNet 101 with three cascading blocks. After ResNet, a modified ASPP is used to compose various scale features. We also use a skip connection to transfer detailed information.

\begin{figure}[H]
  \centering
  \includegraphics[height=13cm,width=0.4\textwidth]{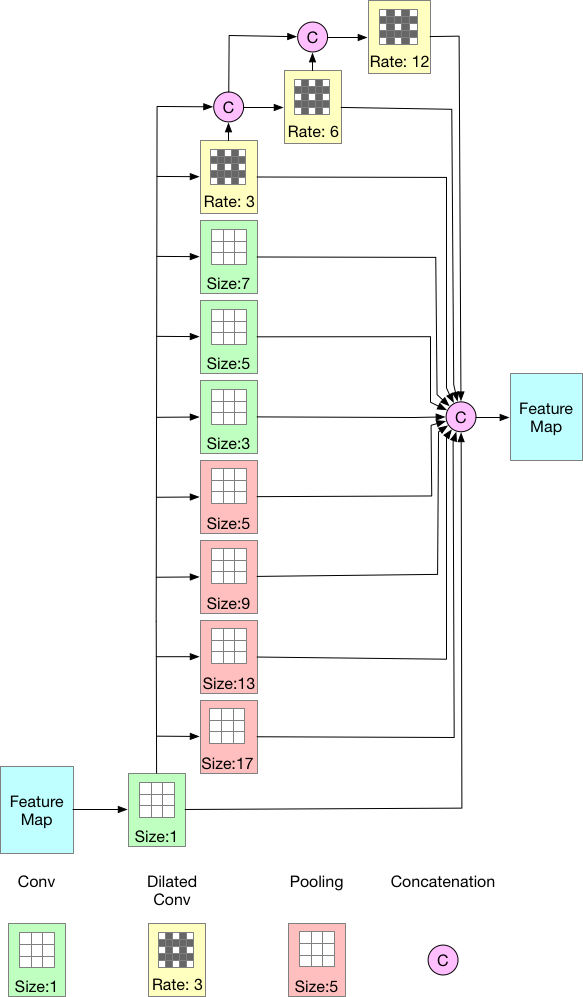}
  \caption{Modified ASPP}
  \label{fig:ASPP}
\end{figure}

Figure \ref{fig:ASPP} shows the structure of the modified ASPP block. A 1x1 convolutional layer is used to compose the feature extracted by ResNet and reduce the number of feature maps. After that, we use a modified ASPP block to extract information in various scales. The modified ASPP has three parts which are dense ASPP, standard convolution layers and pooling layers. Dense ASPP is proposed by \cite{Yang_2018_CVPR} which reduce the influence of margin in ASPP. Considering the vague boundary and low contrast appearance of the lesion, we add standard convolution layers to enhance the ability of neural networks in distinguishing the boundary. The aim of pooling layers is to let the networks consider the surrounding area of the low contrast lesion. The modified ASPP includes three dilated convolutions with rate 3, 6, 12 respectively, three standard convolution layers with size 3, 5, 7 respectively and four pooling layers with size 5, 9, 13, 17 respectively.

\subsection{Pre-processing}
\begin{figure}[H]
  \centering
  \includegraphics[width=0.4\textwidth]{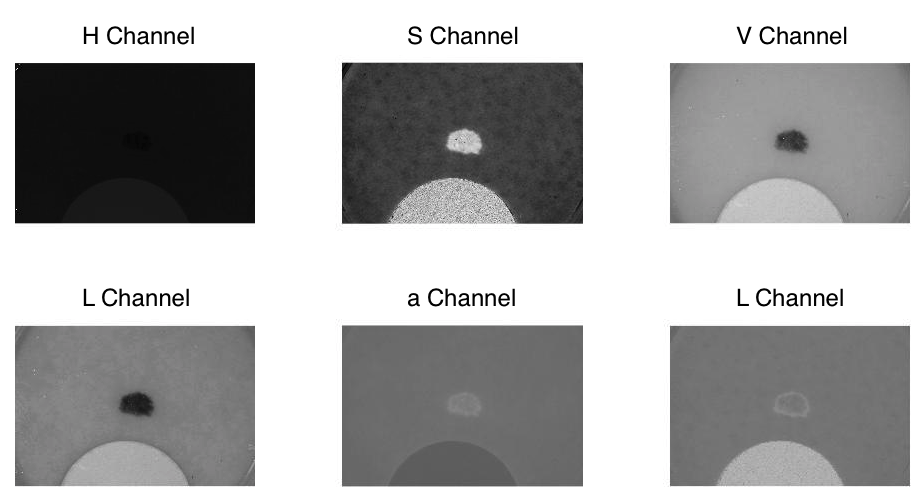}
  \caption{Single channel in HSV and CIELAB colour space}
  \label{fig:hsvlab}
\end{figure}

Instead of only using RBG channels, we combine the SV channels in Hue-Saturation-Value colour space and lab channels in CIELAB space with the RGB channels. These 8 channels are the input of segmentation part. Figure \ref{fig:hsvlab} shows different channels of HSV and CIELAB colour space.

\subsection{Post-processing}
The ensemble is used as the post-processing to increase the performance of segmentation. The input image of the segmentation part is rotated 90, 180 degrees and flipped to generate the other three images. Each image has a result predicted by the segmentation part. The results of the rotated and flipped image need to rotate and flip back to the original image. The final mask is the average of four results of these images.

\begin{figure}[H]
  \centering
  \includegraphics[width=0.5\textwidth]{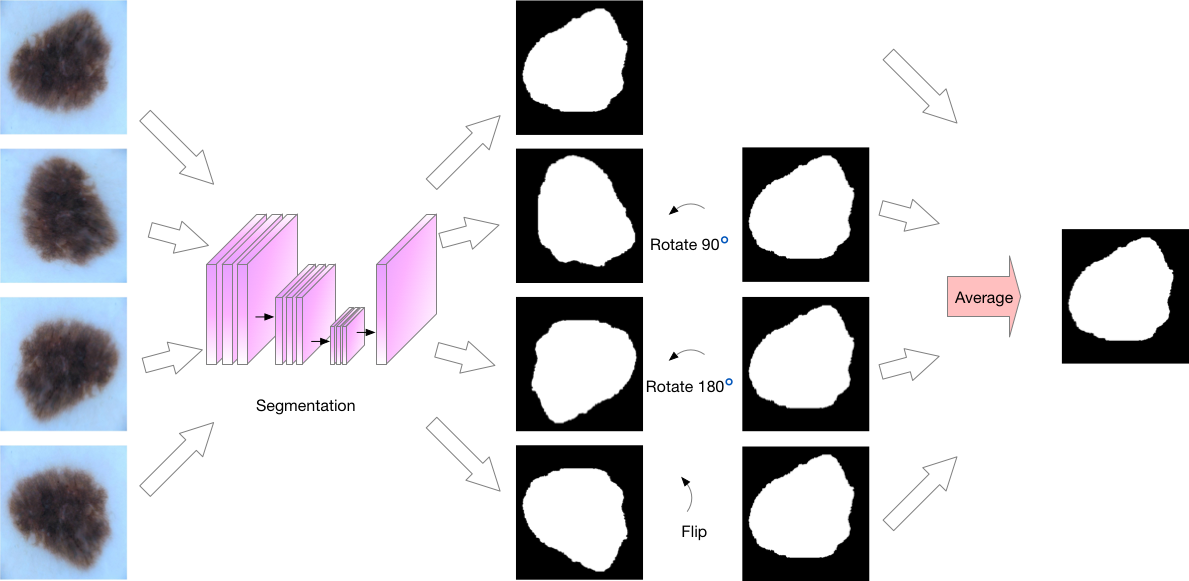}
  \caption{Ensamble}
  \label{fig:Ensamble}
\end{figure}

\subsection{Training}
\begin{figure}[H]
\centering
\begin{subfigure}{.25\textwidth}
  \centering
  \includegraphics[width=0.4\linewidth]{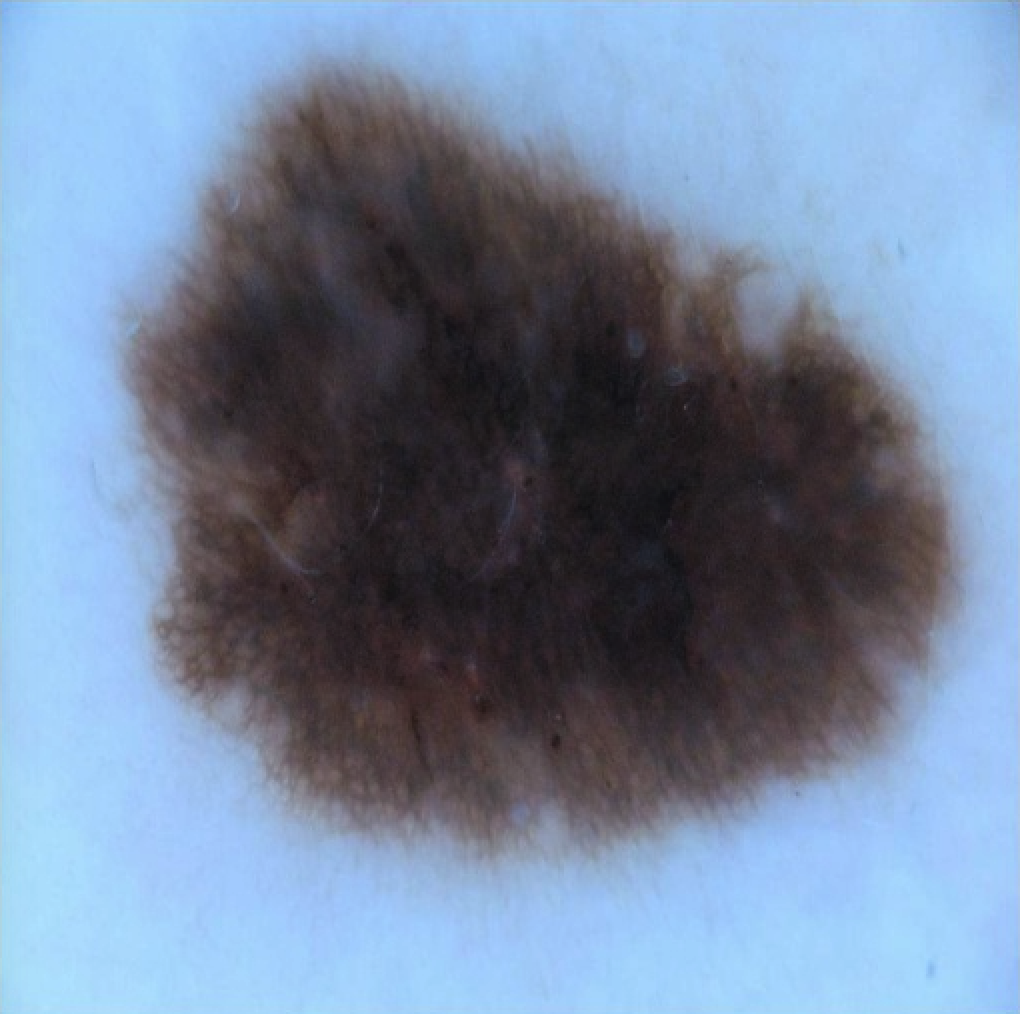}
  \caption{Flip Left to Right}
  \label{fig:aug1-1}
\end{subfigure}%
\begin{subfigure}{.25\textwidth}
  \centering
  \includegraphics[width=0.4\linewidth]{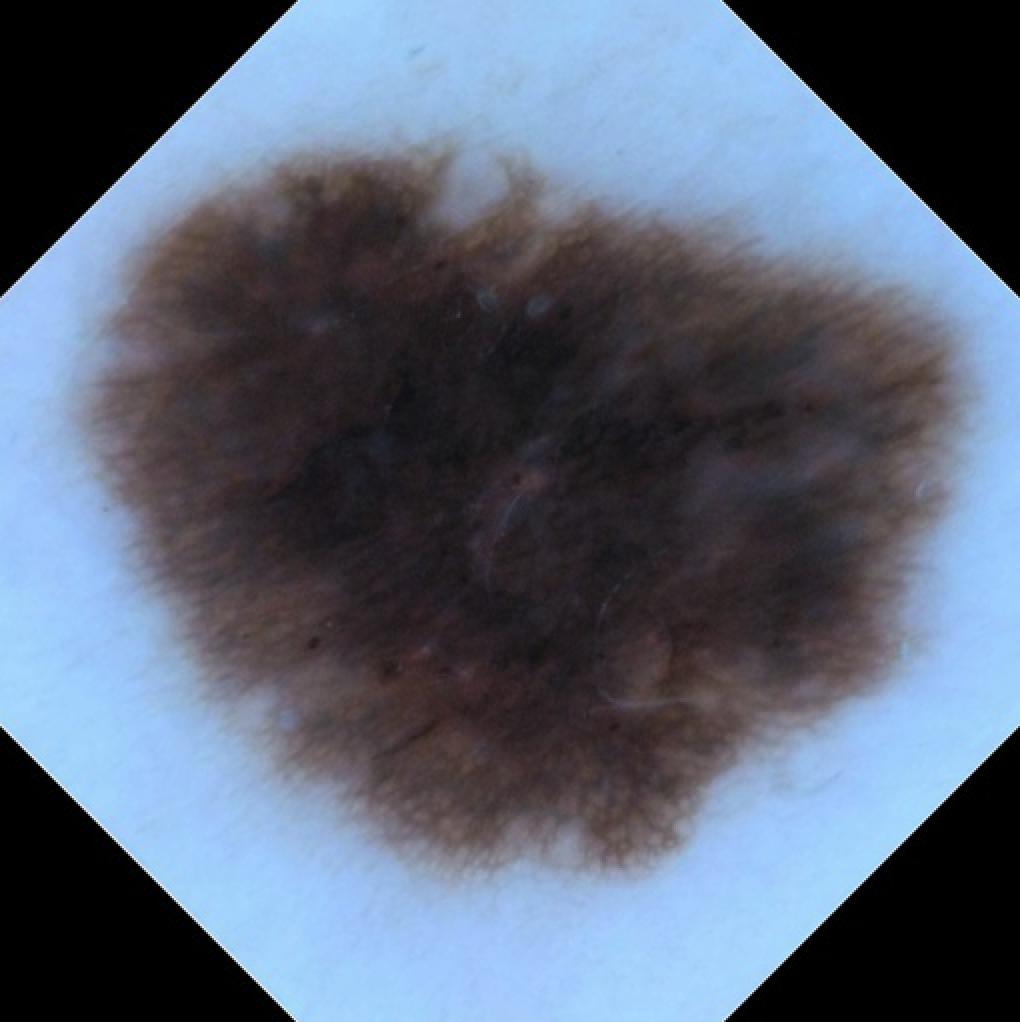}
  \caption{Roatate 45$^\circ$}
  \label{fig:aug1-2}
\end{subfigure}
\caption{Examples for image augmentation}
\label{fig:aug1}
\end{figure}

Image augmentation is used in training of detection and segmentation. Rotation, colour jitter, flip, crop and shear is operated in each image. Each channel of images is scaled to 0 to 1. In segmentation part the size of input images is set to 512x512. Some examples are shown in Figure \ref{fig:aug1}. 

We use the Adam optimiser and set the learning rate to 0.001. The learning rate will be set at 92\% of the previous after each epoch. The batch size is 8. We early stop the training when the net start overfitting. We use the dice loss which is shown in Equation \ref{eq:diceloss}, where $p_{i,j}$ is the prediction in pixel $(i,j)$ and $g_{i,j}$ is the ground truth in pixel $(i,j)$. 

\begin{equation}
\label{eq:diceloss}
L =  -\frac{\sum_{i,j}(p_{i,j} g_{i,j})}{\sum_{i,j}p_{i,j} + \sum_{i,j}g_{i,j}-\sum_{i,j}(p_{i,j} g_{i,j})}
\end{equation}

\begin{figure}[H]
  \centering
  \includegraphics[width=0.3\textwidth]{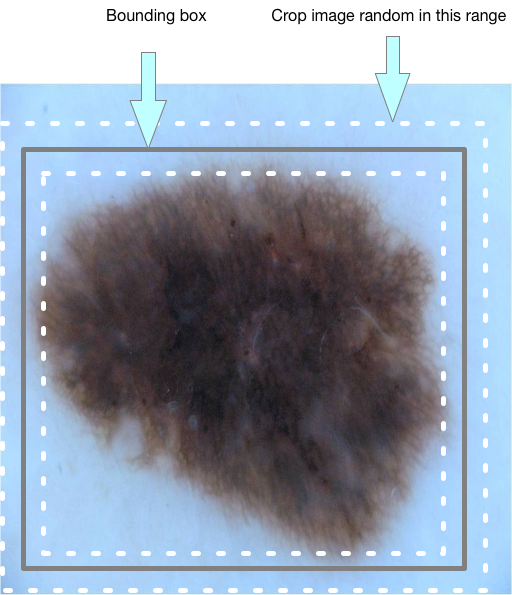}
  \caption{Crop image}
  \label{fig:crop}
\end{figure}

The input image of the segmentation part is crop randomly in a range near the bounding box predicted by detection part. In order to improve the diversity of input images and provide more information about background context. In training, the input image of the segmentation part is cropped from 81\% to 121\% of the bounding box randomly which is shown in Figure \ref{fig:crop}.

\begin{figure*}
 \centering
  \includegraphics[width=0.9\textwidth]{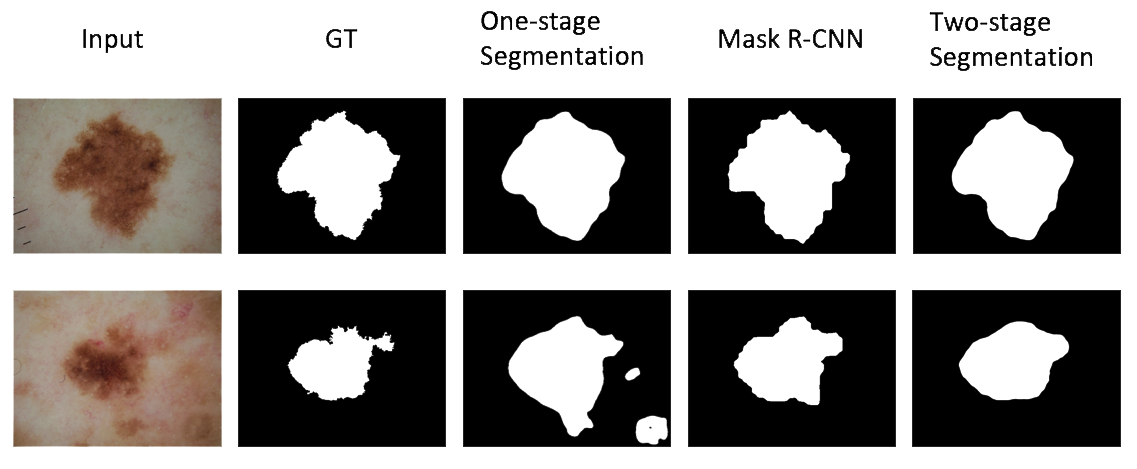}
  \caption{Predicted masks of different segmentation methods}
  \label{fig:result}
\end{figure*}

\subsection{Implementation}
The detection part of our method is implemented by using Pytorch 0.4 in Ubuntu 14.04. The framework is from \url{https://github.com/roytseng-tw/Detectron.pytorch}. The segmentation part is implemented by using Pytorch 0.3.1 in Ubuntu 14.04. The neural networks are trained by two Nvidia 1080 Ti with 60 GB of RAM.

\section{Results}
\begin{table}[H]
\centering
\begin{tabular}{llll}
\hline
Method & Jaccard & Jaccard(>0.65)\\
\hline
Mask R-CNN & 0.825 & 0.787\\
One-stage Segmentation & 0.820 & 0.783\\
Two-stage Segmentation & 0.846 & 0.816\\
\hline
\end{tabular}
\label{table:results}
\caption{Evaluation metrics of different segmentation methods}
\end{table}

The evaluation metrics on 257 images of our two-stage segmentation method with MaskR-CNN and our one-stage segmentation method is shown in Table 1. The thresholded Jaccard of our two-stage method on the official testing set in 0.802. Figure \ref{fig:result} shows the outputs of different segmentation methods. Compared with other methods, The results of our two-stage method has a better location and more smooth edge.

\bibliographystyle{unsrt}
\bibliography{cy_isic_manuscript}

\end{document}